# Intelligent anticipated exploration of web sites

Giovambattista Ianni

Dipartimento di Matematica, Università della Calabria,

87036 Rende (CS), Italy

ianni@deis.unical.it

## Abstract

*In this paper we describe a web search agent, called Global Search Agent (hereafter GSA for short). GSA integrates and enhances several search techniques in order to achieve significant improvements in the user-perceived quality of delivered information as compared to usual web search engines. GSA features intelligent merging of relevant documents from different search engines, anticipated selective exploration and evaluation of links from the current result set, automated derivation of refined queries based on user relevance feedback. System architecture as well as experimental accounts are also illustrated.*

## 1 Introduction

The recent overwhelming proliferation of web-accessible information sources has made the problem of data selection and delivery hard to solve over the Internet.

The everyday use of the Web that one usually carries on with the assistance of web search engines (e.g. Altavista, Lycos, Google [1, 22, 11]) provides good hints to guess the actual difficulty of this problem. Web search engines, that are the usual tools one uses for information gathering, try to face the complexity implied by Internet searching by constructing and maintaining huge indices, which allow relevant document retrieval on a by-keyword basis. Usually indices are built by exhaustively exploring the web by means of "search bots". Search bots do not feature "intelligence", and often are limited to thoroughly explore all the web.

Unfortunately, no existing index successfully tracks all the existing web pages, though some recent project attempted to do so (e.g. the Inktomi indexing service [14]). As a result, the obtained recall capacity (i.e. the fraction of retrieved relevant pages w.r.t. the total of relevant pages) of available web-search engines is far from reaching 100%. Moreover, being based on a generic keyword-based search, ordinary search engines fail to achieve good performances as far as the user-perceived precision (i.e. the fraction of perceived relevant documents w.r.t. the total of retrieved documents) is concerned.

The Global Search Agent, that we illustrate in this paper, tries to improve web search quality on both the recall and the precision sides. As far as the former one is concerned, Global Search enquiries several usual engines and rearrange the set of returned results in such a way to enhance coverage, without diminishing precision. As for the latter measure, GSA implements a user-profile based technique by which in document selection and delivery actual user preferences are suitably taken into account.

Furthermore, GSA features intelligent selection of "promising" paths to explore. This selective approach greatly reduces search space without significantly diminishing precision and/or recall. The anticipated exploration method fruitfully takes advantage of a "locality" principium, stating (roughly) that neighborhood of relevant documents is likely to be rich of relevant documents.

From the realization viewpoint, GSA is a stand-alone application which should be installed within a user Internet-ready machine. From the application, the user can specify its requests as with a usual web-search engine, in the form of a set of keywords.

GSA then queries a relevant set of search engines, hence collects and ranks results from them. The user can browse documents as soon as they are displayed, while the system searches for other results. The search strategy selectively follows links adjacent to the initial ones, to look for further, potentially interesting documents.

As said above, in document selection and scoring, user preferences are taken into account. These are specified by the user in the form of a structured concept tree: the tree structure is employed by GSA to restrict search progress to those documents expected to "match" the specified tree. User preferences are not static, though. Indeed, the user can express his/her approval about documents that GSA returns to him/her. User "opinions" about returned documents are employed to provide the user with suggestions on how to improve the overall correspondence between returned results and user's information needs, also achieving some adaptivity.

GSA allows to schedule periodic searches, to configure the set of queried search engines, the document quality criterion and the aggressiveness of in-depth anticipated search. Furthermore, queries can be delegated to remote instances of the agent, which are able to push back the found results.

The rest of the paper is devoted to discussing, in more detail, the technical features of GSA and the methods implemented therein, focusing, in particular, on anticipated exploration methods. In particular, the next section presents GSA software architecture. Section 3 discusses meta-searching in general. Anticipated exploration and agent pushing via links (using the so called *spiders* – see below) and their management is discussed in Section 4. Setting and managing user preferences are dealt with in Section 5. Section 6 reports an example session, whereas Section 7 accounts for some experiments conducted with GSA in order to establish the effectiveness of the approach in several application contexts. Finally, in Section 8, conclusions are drawn. Related literature and systems are referred to throughout the paper when needed.

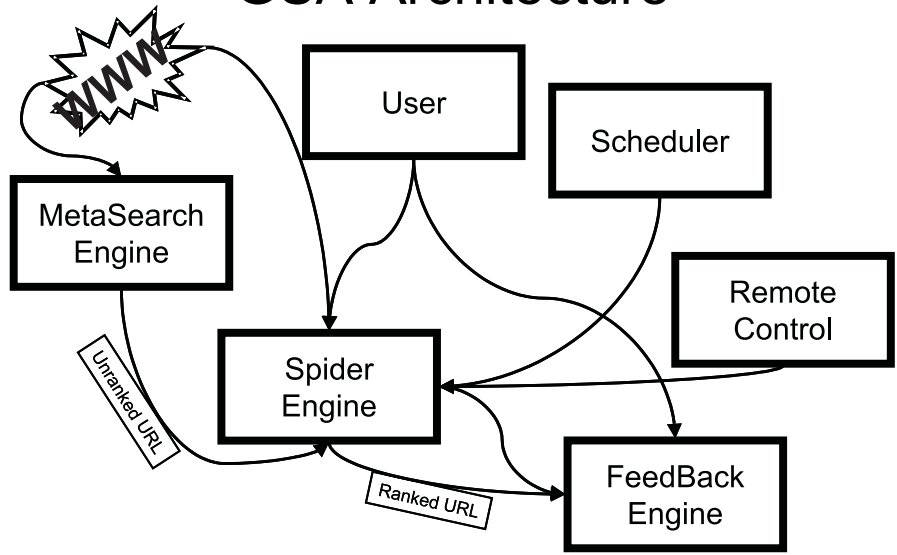

**Figure 1. The GSA architecture**

## 2  System Architecture overview

GSA coordinates and embeds some different methods, conceived in order to achieve effective document retrieval. Figure 1 illustrates GSA architecture where main software modules (which roughly correspond to main system functionalities) are described. Three different "engines" take care of major tasks:

1. The Metasearch Engine (ME in the following), which exploits meta-search on the Web;

2. The Spider Engine (SE in the following), which exploits anticipated exploration;

3. The Feedback Engine (FE in the following), which manages user preferences.

The three engines are shipped within an application which allows to take advantage of these features in order to fruitfully search the Web, and can be activated either (at previously arranged dates) by the Scheduler module, by a Remote Control module, or by the user.

User enters and manages a concept tree containing a set of queries (queries are, as usual, set of keywords), and each query is associated with a set of related documents. The concept tree is conceived in order to maintain a category hierarchy, which each query should be matched to. This structured classification approach assists user in order to handle past searches, and allow to specify user profiles in terms of how and which query are related to any other. Beside this (as it will be clear in the following) the position of a query within the concept tree is helpful to drive search. Moreover, user is able to browse documents and a further feedback can be provided by marking interesting documents.

A query can be "deployed" (meaning that user asks GSA for retrieving pertaining documents), and once this happens, GSA starts SE and ME, which run in parallel.

The purpose of ME is to retrieve interesting documents from a given set of web search engines. ME submit a given query to a set of web search engines; each search engine is enquired through a wrapper (i.e. a specialized driver). Each wrapper is able to query the corresponding search engine in the correct way, and to select significant URLs from the resulting, semi-structured, marked-up page. In other words, a wrapper purges retrieved answer pages from noisy hyperlinks (e.g. advertising links), providing a structured list of URLs. As soon as they are collected, URLs are then proposed to SE.

SE manages a collection of "spiders", which are employed in order to retrieve, validate and score documents. A spider is an agent which crawls the web pursuing the objective of finding interesting documents.

Whenever some starting document is proposed to SE (usually starting URLs come either from ME or from the user/scheduler/remote control input), spiders intelligently follow the hyperlink chain, deciding which documents are to be presented to the user, and which directions are worth to be explored. In order to quantify interestingness of a document, GSA employs a suitable scoring function, which takes into account similarity, frequency and mutual distance between entered keywords, with respect a given document.

Decision regarding which directions are suitable for further exploration are taken by giving each spider memory of past visited documents, and by means of a "locality" principium (i.e. "high quality documents are often surrounded by further high quality documents"). Spiders increase likelihood of further exploration whenever relevant documents were recently explored. They tend, instead, to "die" if no relevant documents were found in the past. Score of the document with respect to the position of the given query within the concept tree is considered as a further relevance criterion.

Evaluation may be explicitly stopped at any moment, although it is allowed to analyze results while the system is still performing the search. In any case, search is performed in a finite time.

The Feedback Engine works off-line. User can specify his/her opinion about which of the returned documents are interesting and which are uninteresting, marking entries of the document list accordingly. Once the user opinion is given (even on a small subset of the overall set of retrieved documents), FE can be started. The output of FE is a set of relevant words, suggested to the user in order to refine the search. User can employ suggested words to fruitfully modify the concept tree to which the current query belongs to. The chosen strategy of FE consists in the selecting, from a restricted set of candidate terms, words which significatively improve scoring gap between user-perceived interesting documents and uninteresting ones.

## 3   Meta-searching

The merging of documents found by different search engines enhances the overall web coverage and the quality of documents found: several automatic collection techniques from different search engines are known, such as the ones used in MetaCrawler, Profusion, Inquirus, SavvySearch [26, 10, 16, 6] and the ones from the recent commercial

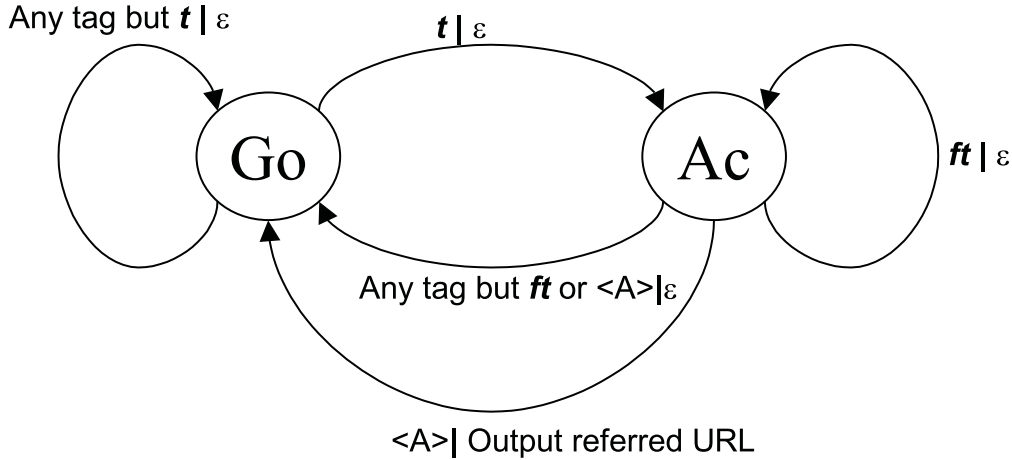

**Figure 2. The simple wrapper automaton $\mathcal{W}$.**

experience of WebFerret, Copernic and Lexibot [20, 5, 19].

However, several problems are to be addressed when the need of successfully merge results from heterogenous search resources arise. A first category of issues to be addressed deals with the semistructured way informations are passed to and from informative sources. First, the right way to query each search engine is very different from one another. Second, results are sent back to the user in a semi-structured form (usually an HTML page) containing a list of proposed URLs, including page title, and, usually, a short abstract of the document: an ad hoc wrapper is then needed, in order to extract information for each different search resource the agent may desire to query. Each wrapper acts as an independent entity and supplies the application with a set of results (given in an engine-independent form) when a new query is submitted to it. Third, lifetime of each wrapper may be very short, since slight modifications in the way a result page is encoded can affect considerably wrapper precision and recall.

In other words, wrappers act as very specific transducing automata, which expect to find information within very specific contexts in HTML (or any marked up document) pages. Since the way in which data are presented (and the relative encoding) changes periodically at a relatively high rate, wrappers are to be redesigned at the same rate as well, in order to fit with new data presentation structures.

This problem becomes crucial if the number of resources to be consulted is very large. Assuming that a specific resource changes its data presentation structure once a year (an optimistic assumption), a meta-search engine able to reach $600$ or more search engines (such as Lexibot [19]) forces its maintainers to rewrite (or regenerate) about two wrappers per day. Such issues are usually addressed by providing periodic updates of the featured wrapper library.

GSA wrappers are simple finite state automata dealing with marked up documents. For example, a wrapper automaton template $\mathcal{W}$ is described in Figure 2. **t** and **ft** are given tags, whereas <A> is the traditional anchor tag. The input stream is assumed to be a sequence of tags belonging to some mark up language. Notation for state transitions is given as usual in the form *input|action*, where $\epsilon$ means "do nothing". Starting state is labelled as "Go".

It can be noted that $\mathcal{W}$ repeatedly and selectively accepts URLs whenever they are immediately preceded by the tag **t** (possibly preceded by some sequence of **ft** occurrences). Indeed, this corresponds to recognizing substrings of the document at hand belonging to the regular language $W \equiv \mathbf{t}\,\mathbf{ft}^*\mathtt{<A>}$.

At the moment, eleven out of twelve wrappers of the GSA collection are fully operative using the simple structure shown above. Wrapper maintenance burden is very light since in many cases, automata which cease to be operative are fixed by changing **t** and **ft** values. Values for Lycos [22] wrapper are, for instance[1], `<LI>` and `<FONT>`, for **t** and **ft**, respectively.

It is known that finite state automata are often powerful enough to extract information from web sources, and rarely more powerful models such as stack automata are needed [23]. Thus, we argue that the employment of very simple classes of finite state automata could be a general trend in this specialized case, although it is topic of future research to better investigate on properties of search engines page encoding in order to assist automatic wrapper generation ([3, 23]).

Anyway, this approach simplifies the task of designing wrappers, which, furthermore, results to be light-weighted, less sensitive to modifications within the source structure, and thus long lasting.

A second category of issues deals with the problem of suitably merging results coming from different sources. In order to score documents, each web source introduces its own relevance criterion. Furthermore, identical documents may be referenced from several sources.

[10, 12] and [6], address the issue by attempting to merge results using heuristic techniques intended to deal with the unknown ranking criterion of each search engine. In order to give to a document a suitable score, these techniques try to take into account how several different search engines present each document, which is the most affordable search engine in terms of time response and precision for the given query, and, if necessary, abstract texts usually proposed by search engines are interpreted and provided in terms of score given to documents.

However, this approach did not prove to be clearly useful in order to provide a suitable sorting of retrieved documents; moreover it would force GSA to gather all the results before a single document can be displayed. Thus, wrappers do not provide neither relevance values nor other data different from a pure list of candidate URLs; the ranking, extraction (of additional information such as titles and abstract) and merging steps are deferred to following phases.

Table 1 summarizes peculiar technical choices of GSA Metasearch Engine with respect to the three commercial meta-search products Copernic, Web Ferret and Lexibot [5, 20, 19].

As far as wrapper updating is concerned, the modular architecture of the ME allows, if needed, rapid delivery of up to date wrappers. Commercial products feature sometimes automatic update services (Copernic), manual updates (Lexibot), or no update service at all (Webferret). ME wrappers are not in charge of measuring quality of URLs, since this task is delegated to a postponed step featuring direct retrieval of documents. Copernic is the only application which features quality measurement based on heuristics eliminating the need of direct document retrieval, whereas Webferret does not consider this issue at all. As far as auxiliary information extraction (abstracts, titles, etc.) is concerned, GSA ME prefer to defer this task to direct retrieval steps, whereas other products (Copernic, Webferret) feature immediate extraction. Lexibot does not address this issue at all.

---

[1] we refer here to the page presentation encoding of Lycos in July 2001.

|  | **GSA ME** | **Copernic 2001** | **Web Ferret 4.0** | **Lexibot 2.0** |
|---|---|---|---|---|
| Wrappers | Automatic update ready | Periodic automatic update | No update | Update on request |
| Scoring | Deferred | Immediate | No | Deferred |
| Auxiliary information extraction | Deferred | Immediate | Immediate | No |

**Table 1. Differences between GSA, Copernic, Web Ferret and Lexibot .**

## 4 Web-crawling and adaptive exploration

When the Spider Engine is prompted for a new, potentially relevant, result, a new entity, called *spider* is created and deployed. Spiders are able to create and deploy further generations of child spiders and act mostly in parallel. For simplicity of exposition, we present here a partially sequentialized version of the algorithm which encodes the behavior of each spider (Figure 3).

A spider retrieves the document on which it is started on, establishes its score and its *combined score* (Lines 3 and 4), and decides

- if it is worth to show the current document to the user (Lines 5 and 6);

- if it is worth to pursue the task of following the links included in the current document (Line 7 and following ones).

In the former case, the current document is scored by a suitable ranking criterion (see Subsection 4.1 for details), and then it is displayed if the achieved score exceeds a given threshold provided that the document at hand was not displayed before (Line 5).

In the latter case a new child spider is started, one for each link found (Line 9). The decision of launching new spiders on following documents takes in consideration:

1. that the depth of the originating spider (i.e. the distance of the current document from the document the search originated) does not exceed a given threshold (Line 7). Although termination of the search is theoretically guaranteed (see Theorem 4.9), it is not affordable to explore all the Web, hence this constraint is introduced in order to control proliferation of spiders, and, in every case, guarantee termination of the search in a reasonable amount of time;

2. the *happiness* value exceed a given threshold (Line 8). The happiness function, described in Subsection 4.2, measures the quality of the last documents scored by

```
Input: a spider spider started on a URL url, a set of words w;

begin
(1)   if url is not a feasible document or it is not reachable then
(2)       stop this thread;
(3)   s = rank(url,w);
(4)   c = combinedscore(url,w);
(5)   if ( s > DISPLAYTHRESHOLD ) and (url was never presented to the user) then
(6)       present url to the user;
(7)   if (spider.depth < MAXDEPTH)
(8)       and (spider.happiness() > HAPPINESSTHRESHOLD) then
(9)           for each URL u pointed by url do
(10)              if (u was never visited by a spider) then
(11)                  schedule a new spider on u with depth set to spider.depth+1,
(12)                  history = spider.history ∪⟨spider.depth, c⟩, and
(13)                  priority proportional to spider.happiness();
end;
```

**Figure 3. Algorithm which controls the behavior of a spider**

the generation of spiders the current one belongs to; if happiness decreases under a given threshold, the current spider "dies", and no further spiders are generated, whereas spiders with high happiness value are scheduled with higher priority and early deployed (Line 13);

3. the document at hand was never analyzed, has a valid format and is reachable (Lines 1,2 and 10). Note that feasibleness and reachability of a document are established a posteriori (after scheduling) by child spiders[2] (Lines 1 and 2).

Child spiders inherit, from the originating spider, history of latest visited documents, in the form of a set of couples $\langle d, c \rangle$. Each pair $\langle d, c \rangle$ tracks that the generation of spiders, which the current one belongs to, visited a document with combined score $c$, and visit was performed by a spider at depth $d$. Moreover, child spiders are initialized to a depth incremented by one with respect to the originating spider (Line 11). Combined score (see Definition 4.4) is a measure of the quality of the document at hand with respect to the entire concept tree which the evaluated query belongs to.

This approach sacrifices some efficiency, since each document must be retrieved. However, it provides effective removal of not well ranked and/or not reachable documents, and allows homogenous merging of results coming from different search engine sources. The number of analyzed documents is effectively reduced by employing a pruning criterion based on the locality principium which leads to the introduction of the happiness concept (see Subsection 4.2).

---
[2]Reachability is established at operating system level by internal timeout thresholds.

### 4.1 The ranking function

In order to associate a ranking value to a document, we adapted a ranking function based on the one proposed in [16]. This function embeds three components: *a)* a presence component, whose value is proportional to the presence of almost one *significant* occurrence of a given term within the page text, *b)* a frequency component, that weighs the overall quantity of significant occurrences for a given term within a page, and *c)* a distance component, that weighs the overall distance between significant occurrences of the given terms.

The concept of significance for a word occurrence is often introduced through a *stemming* algorithm. Two words are considered identical whenever their *stem* matches (e.g. "opener" and "opening" have same stem). Word stemming algorithms are commonly specialized on the particular lexicon the text is assumed to be written in.

Roughly speaking, in GSA an occurrence of a word $W_2$ is considered *significant* if it is very similar to a user-entered keyword. Thus, a concept of similarity is introduced in order to consider the stem of each word when this has to be compared with another one: differently from the classical Porter's algorithm (see [9]), we chose a stemming criterion independent on the language that the text is supposed to be written in.

Given a word $w_j$ (resp. a set $S$), let $|w_j|$ be the length of $w_j$ in characters (resp. let $|S|$ be the cardinality of $S$).

**Definition 4.1** The similarity function $sim$ takes two words $w_1$ and $w_2$ and returns a value between 0 and 1. It returns a value near to 1 if $w_1$ and $w_2$ are very similar (i.e. the first one can be considered a *significant* occurrence of the second one), and a smaller in the opposite circumstance. In particular, $sim$ is defined as

$$sim(w_1, w_2) = \left(\frac{|x|}{|w_1|}\right)^4$$

where $x$ is the common stem between $w_1$ and $w_2$ (e.g. the common stem of "opener" and "opening" is "open"). □

If the length of $w_1$ and $w_2$ are not the same, similarity is weighed w.r.t. the first string, which usually is a user-entered keyword. Note that, in general, $sim(a,b) \neq sim(b,a)$. The ratio is to the 4th power in order to serve the purpose of lowering similarity for words shorter than the first argument word. As an example $sim("java", "javadoc") = 1$, whereas $sim("java", "jav") = 0.31$. A suitable threshold value, to be compared with similarity values, is employed in order to cut off insignificant occurrences.

Here, and in the following, we assume that a document $D$ is a sequence of words (purged from mark-up strings of any sort).

**Definition 4.2** The ranking function $rank$ takes a document $D$ and a set $W$ of given keywords (the search query to be considered), and is evaluated as shown in Figure 4. In detail:

$$\begin{aligned}
rank(D, W) &= k_0 \frac{f}{f + k_1}, \text{where} \\
f &= b_0 + f_0 + d_0, \\
b_0 &= k_2 \frac{N_p}{|W|} \\
f_0 &= k_3 \frac{N_t}{|W|} \\
d_0 &= k_4 \frac{k_5 - \frac{\sum_{i=1}^{N_s-1} \sum_{j=i+1}^{N_s} \min(d(i,j), k_5)}{\sum_{k=0}^{N_s-1}(N_p - k)}}{k_5}
\end{aligned}$$

**Figure 4. The *rank* function**

- The term $b_0$ represents the presence component: the value

$$N_p = \sum_{\substack{w \in W \\ sim(w,d) > T_s}} \max_{d \in D} sim(w, d)$$

is the sum of *presence* values (exceeding the threshold $T_s$) of each term. Given a word $w \in W$, the *presence* is the best similarity found for $w$ with respect to all the words of $D$ (i.e. the best significative occurrence of $w$ within $D$).

- The term $f_0$ represents the frequency component: the value

$$N_t = \sum_{w \in W} \sum_{j=1}^{\lfloor H \rfloor} \frac{1}{2^j}, \text{ where}$$

$$H = \sum_{\substack{d \in D \\ sim(w,d) > T_s}} sim(w, d)$$

is the total sum of significant occurrences of the words of $W$ within $D$; each significant occurrence is weighed by the corresponding similarity value. Note that the more a word occurs in a document the less further occurrences are weighed.

- The term $d_0$ represents the distance component: the value $N_s$ is the number of words of $W$ with almost a significative occurrence within $d$. If we denote by $w_i$ the $i$-th element of $W$, then $d(i,j)$ represents the minimum distance found between two significant occurrences of the words $w_i$ and $w_j$;

- $k_0$ and $k_1$ are two constants controlling, respectively, amplitude and asymptotical trend of the $rank$ function, whereas $k_2, k_3, k_4$ are suitably chosen weights for each of the three components, and $k_5$ is the maximum distance (in words) to be considered significant between two significative occurrences of terms in $W$.

The current prototype allows these latter values to be set by the user to desired values. However, as we experimentally appreciated, methods which privilege distance and presence components showed to be effective (see, e.g. [4]), thus chosen values throughout this paper are set in order to further penalize frequency component.

Given a set of query words $W$, a document $D$ is displayed whenever $rank(W, D)$ exceeds the given threshold.

Differently from [16], our ranking function *(a)* embeds directly some stemming techniques, *(b)* naturally penalizes frequency component, *(c)* expresses distances in words and not in characters and, *(d)* is bounded within a given range (in fact, $rank$ ranges from $0$ to the asymptotic value $k_0$). This latter feature eliminates the need of normalizing the rank values at the end of the search and the need of knowing a priori the maximum and minimum score of retrieved documents, providing a sort of fixed relevance scale whose maximum value tends to the right edge (i.e. $k_0$) of the allowed score interval.

### 4.2 The concept tree, the happiness function and the locality principium

The idea of "happiness" associated to a spider couples the approach of Letizia, and Webwatcher [21, 15], with some "intelligent" criterion of exploration, allowing effective pruning on the number of visited pages, and rapid recover of interesting documents. While Letizia agents are guided by user implicit preferences (e.g. times a URL has been clicked, time user has spent on a given page), to early detect the direction an agent should follow, GSA spiders can decide directions to follow pretty independently on the user and far and far away from the pages the user is considering at the moment.

By a locality principium, pages related to a given topic should be surrounded by several further pages concerning the same topic; thus, it makes sense to guide exploration by taking into account this idea, in order to tailor search space by excluding paths not leading to high quality documents. Happiness make each spider able to take into account a) the list of the ranking values scored by the past visited documents, and b) the *concept subtree*, which the originating query refers to.

**Definition 4.3** A *concept tree* is a tree $T = \langle N, E \rangle$ rooted at a given node $r$, where each node $n \in N$ is a set of words, and each element $e \in E$ is a pair $\langle u, v \rangle$ where $u \in N$ and $v \in N$. Leaf nodes of $T$ will represent a user *query*, whereas intermediate nodes will be called *concepts*. □

In general, GSA users might enter concept forests (i.e. a set of different concept trees). We will assume, without loss of generality, that, whenever a forest of concepts $F$ has to be dealt with, a dummy root concept $r = \{\}$, is introduced in order to bind together separate components of $F$.

**Definition 4.4** It is given a sequence $L = \{d_1, \ldots, d_n\}$ (where $d_1$ represents the last visited document, and $d_n$ the last but $n - 1$) of recently visited documents, a concept tree $T = \langle N, E \rangle$ and a query $q_0 \in N$.

Let $q_1, \ldots, q_m$ be concepts in $N$ appearing on the unique path from $q_0$ to the root of $T$. The *combined score* $S(d, q_0)$ for a given document $d$ is

$$S(d, q_0) = \frac{rank(d, q_z)}{(z+1)k_6^{\min(1,z)}}$$

where $z$ is

$$z = \min_{j=0..m} j \text{ such that } \left[ \frac{rank(d, q_j)}{(j+1)k_6^{\min(1,j)}} > k_7 \right]$$

and $k_6$ and $k_7$ are two fixed parameters. Note that $m$ can be eventually set to zero (i.e. a query may have no ancestor concepts). □

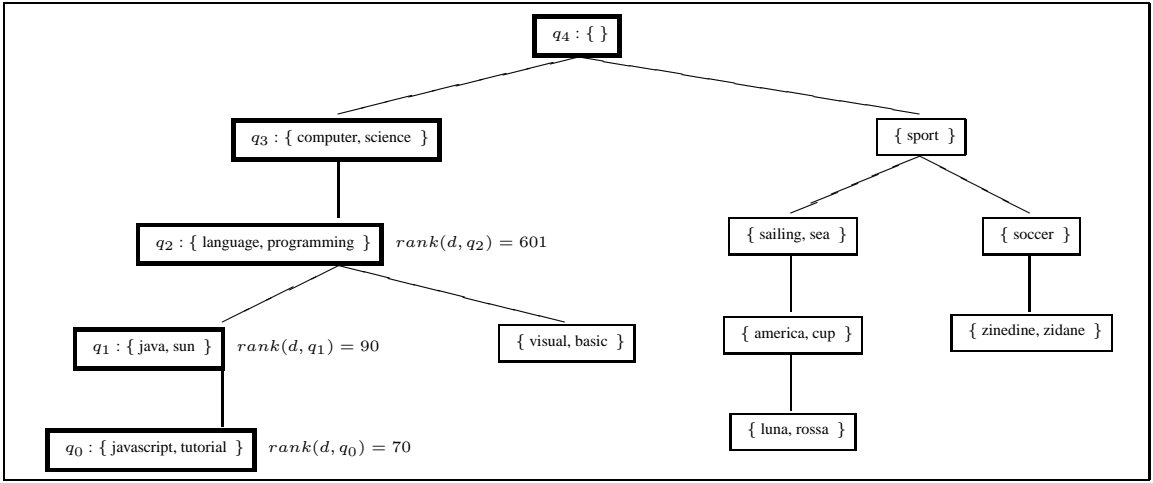

**Figure 5. An example of evaluation on a concept tree for a given document** $d$

Concepts are set of words which are considered to be more general, with respect to keyword entered in a search query. In other terms, assume a query $q$, and an ancestor concept $c$ of $q$ is entered. Roughly speaking, whenever a document $d$ cannot achieve a sufficient value for $rank(d, q)$, the notion of combined score gives "further chances" to the document at hand, by considering the value of $rank(d, c)$. However, the importance of scores computed using concepts' words is lighter the farther a concept is from the original query.

As an example (see Figure 5), the words "language" and "programming", will have less importance with respect to "java" and "sun", if the query $q_0 = \{$"javascript","tutorial"$\}$ is submitted against a given document $d$. However, assuming $k_6 = 2$ and $k_7 = 100$ then $S(d, q_0) = \frac{rank(d,q_2)}{3*2^1} = 100.16$, since $rank(d, q_1)$ and $rank(d, q_2)$ are too low in order to give a value exceeding $k_7$ for $S(d, q_0)$. Indeed, $\frac{rank(d,q_0)}{1*2^0} = 70$ and $\frac{rank(d,q_1)}{2*2^1} = 22.5$.

We are now ready to discuss how the happiness function comes into play.

**Definition 4.5** It is given a set of $n$ documents $L$, and a query $q_0$ belonging to a concept tree $T$. The *happiness* $h$ is defined as:

$$h(L, T, q_0) = \sum_{d \in L} \frac{S(d, q_0)}{n}$$

where each term $S(d, q_0)$ is the combined score of the document $d$, given the set of documents $L$, the concept tree $T$ and the query $q_0$. □

Roughly speaking, happiness is a value similar to the average score of the last $n$ documents visited by the generation of spiders at hand. However, combined score is employed instead of *rank*. When a document scores too a low *rank* value (non exceeding $k_7$), spiders try to score it using the antecedent node of the original query, and so forth, until the root node is reached or a worth score is obtained. The combined score gives decreasing weight (decreasing factor is controlled by the value of $k_6$) to words belonging to concepts which are far from the original query node. Thus, happiness is designed in order to encourage with high priority those spiders which recently explored documents of considerable quality with respect to the topic at hand. However, some chance is also given to spiders which recently explored documents regarding more general subjects (i.e. topics identified by ancestor concepts of the subject at hand).

### 4.3 Exploration termination

It is worth to note that Algorithm of Figure 3 terminates in finite time. Indeed, this algorithm is formally equivalent to a particular way of visiting a labeled graph.

**Definition 4.6** A *webgraph* is a triple $\langle N, E, r \rangle$ where $N$ is a set of nodes, $E$ is a binary relation such that $E \subseteq N \times N$, and $r$ is a function s.t. $r : N \mapsto [0, 1]$. □

**Definition 4.7** It is given a webgraph $G = \langle N, E, r \rangle$ and an integer $m$. Let $p = \langle p_1, \ldots, p_n \rangle$ be a path on $G$. Define $h_m$ as

$$h_m(p) = \sum_{i=1}^{min(m,n)} \frac{r(p_i)}{min(m,n)}$$

The meaning of a webgraph $G = \langle N, E, r \rangle$ is intended in order to represent the web as a graph, where $N$ is a set of documents, linked together by a set $E$ of (direct) edges; each node $n \in N$ is coupled with its weight $r(N)$, representing a measure of its quality (i.e. its $rank$ value). The function $h_m$ will represent a generalization of the happiness function. We assume here the *quality* of a node is measured by means of a rational whose value is normalized between 0 and 1.

Algorithm of Figure 3 may be encoded more formally as follows.

**Algorithm 4.8** *Input*: a webgraph $G = \langle N, E, r \rangle$, a starting node $u \in N$, two rational values $ht$ and $dt$, an integer $m > 0$.
*Data structures*: a priority queue $Q$, consisting of a set of pairs $\langle s, n \rangle$ where $s$ is a queue of at most $m$ nodes and $n$ is an element of $N$.
*Output*: a set $M \subseteq N$.

(1)  put $\langle \langle u \rangle, u \rangle$ in Q;
(2)  **while** Q is not empty **do begin**
(3)      extract from $Q$ a pair $\langle s, n \rangle$ s.t. $s = \langle s_1, \ldots, s_h \rangle$ and $h_m(s)$ is maximum.
(4)      **if** $(r(n) > dt)$ **then**
(5)          output $n$;
(6)      **if** $(h_m(s) > ht)$ **then**
(7)          **for each** $v \in N$ s.t. $\langle n, v \rangle \in E$ **do**
(8)              **if** ($v$ is not marked as visited) **then begin**
(9)                  mark $v$ as visited;
(10)                 **if** $h = m$ **then**
(11)                     put in $Q$ a pair $\langle \langle s_2, \ldots, s_m, v \rangle, v \rangle$
(12)                 **else**
(13)                     put in $Q$ a pair $\langle \langle s_1, \ldots, s_h, v \rangle, v \rangle$
(14)             **end**;
(15) **end**;

As it can be seen, this algorithm performs a usual visit of a given webgraph $G = \langle N, E, r \rangle$, but:

- If a node $n_m$ is reached through a path $\langle n_1, \ldots, n_m \rangle$, $n_i \in N (1 \leq i \leq m)$ such that $h_m(\langle n_1, \ldots, n_m \rangle) \leq ht$ (a happiness threshold), following nodes will not be visited;

- Nodes $n \in N$ s.t. $r(n) > dt$ (a display threshold) will be output;

Note that Algorithm 4.8 does not introduces a constraint on the maximum distance reachable from the originating node. Indeed, assuming $G$ is finite ( though it can be as huge as the Web ), this constraint is not necessary in order to guarantee termination of the visit (although, of course, this latter constraint is necessary to guarantee termination in a reasonable amount time).

**Theorem 4.9** *Algorithm 4.8 terminates in finite time.*

**Proof.** (Sketch). Just observe that since Algorithm 4.8 performs a visit of a webgraph $G$, each iteration causes some node to be marked, and no node can be enqueued for visit if it was already visited. □

One might argue if it is worth to visit twice or more times a document which was earlier explored, but that it has been reached from a spider with greater happiness than the spider which carried out the first visit of the document. This would allow potentially "luckier" spiders to discover new promising paths to follow within the link structure. However, Algorithm 4.8 allows a node to be visited only once. We can show this does not guarantee all the "promising" paths are visited.

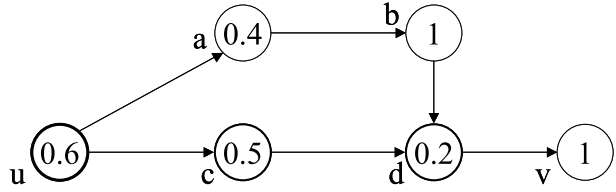

**Figure 6. Counterexample for Theorem 4.11.**

**Definition 4.10** It is given a webgraph $G = \langle N, E, r \rangle$, a starting node $u \in N$, an ending node $v \in N$, an integer $m$, and a rational $ht$. A *promising path* $p = \langle u, p_1, \ldots, p_h, v \rangle$, where $p_1, \ldots, p_h \in N$, is such that $p$ has no sub-paths $s_p = \langle a_1, \ldots, a_m \rangle$ such that

$$h_m(s_p) \leq ht,$$

and for each $t, 0 \leq t < min(m, h)$, no path of the form $\langle u, p_1, \ldots, p_t \rangle$ is such that

$$h_m(\langle u, p_1, \ldots, p_t \rangle) \leq ht$$

□

As an example, assume $ht = 0.5$ and $m = 3$. The path $u, p_1, p_2, p_3, v$ is promising if

- $\left( \frac{r(u)+r(p_1)+r(p_2)}{3} > 0.5 \right) \wedge \left( \frac{r(p_1)+r(p_2)+r(p_3)}{3} > 0.5 \right) \wedge \left( \frac{r(p_2)+r(p_3)+r(v)}{3} > 0.5 \right)$, and

- $(r(u) > 0.5) \wedge \left( \frac{r(u)+r(p_1)}{2} > 0.5 \right)$;

**Theorem 4.11** *It is given a webgraph $G = \langle N, E, r \rangle$, a starting node $u \in N$, a rational $ht$ and an integer $m$. Algorithm 4.8 does not reach all the nodes $v \in N$ for which there exists a promising path from $u$ to $v$.*

**Proof.** (Sketch). A simple counterexample is given in Figure 6. Assume $ht = 0.4$ and $m = 2$. Given a node $n \in N$ the value $r(n)$ is indicated inside the graphical representation of $n$. It can be noted that the path $p_1 = \langle u, a, b, d, v \rangle$ is promising, whereas path $p_2 = \langle u, c, d, v \rangle$ is not. However, node $d$ will be enqueued (and marked as visited) within the queue $Q$ through the pair $\langle \langle c, d \rangle, d \rangle$, since the node $c$ will have greater priority than $a$. Indeed, $c$ will be enqueued through the pair $\langle \langle u, c \rangle, d \rangle$, whereas $a$ will be enqueued through the pair $\langle \langle u, a \rangle, d \rangle$. Thus, $v$ will not be enqueued since $h_m(\langle c, d \rangle) = \frac{0.5+0.2}{2}$ does not exceed $0.4$. □

It turns out that it is possible to avoid promising paths to be discarded by allowing multiple visit of the same node. The following algorithm meets these requirements, and terminates in a finite amount of time, although there are some practical concerns to be discussed.

**Algorithm 4.12** *Input*: a webgraph $G = \langle N, E, r\rangle$, a starting node $u \in N$, two rational values $ht$ and $dt$, an integer $m > 0$.
*Data structures*: a priority queue $Q$, consisting of a set of pairs $\langle s, n\rangle$ where $s$ is a queue of at most $m$ rational values and $n$ is an element of $N$. A map $M$ constituted of associations $M(n) \mapsto h$ where $n \in N$ and $h$ is a rational number (Initially each node $n \in N$ is mapped to 0).
*Output*: a set $M \subseteq N$.

(1)  put $\langle\langle u\rangle, u\rangle$ in Q;
(2)  **while** Q is not empty **do begin**
(3)    extract a pair $\langle s, n\rangle$ from $Q$, where $s = \langle s_1, \ldots, s_h\rangle$ and $h_m(s)$ is maximum.
(4)    **if** $(r(n) > dt)$ **then**
(5)      output $n$;
(6)    **if** $(h_m(s) > ht)$ **then**
(7)      **for each** $v \in N$ s.t. $\langle n, v\rangle \in E$ **do**
(8)        **if** $(M(v) < h_m(s))$ **then begin**
(9)          set $M(v) = h_m(s)$;
(10)         **if** $h = m$ **then**
(11)           put in $Q$ a pair $\langle\langle s_2, \ldots, s_m, r(v)\rangle, v\rangle$
(12)         **else**
(13)           put in $Q$ a pair $\langle\langle s_1, \ldots, s_h, r(v)\rangle, v\rangle$
(14)       **end**;
(15) **end**;

This algorithm does not explicitly mark nodes: a node is visited again if it was past explored through a path (i.e. a spider) with an happiness value lower than the current one. Next, we prove that Algorithm 4.12 terminates in a finite time.

**Theorem 4.13** *It is given a webgraph $G = \langle N, E, r\rangle$, a starting node $u \in N$, a rational $ht$ and an integer $m$. Algorithm 4.12 terminates in a finite time.*

**Proof.** (Sketch). Let $v \in N$. We show that $v$ can be visited only a finite number of times. For simplicity, we assume distance from $u$ to $v$ is at least $m + 1$. Proof can be easily extended to the general case.

Let $P$ be the set of paths of the form $\langle p_1, \ldots, p_m\rangle$, where $\langle p_m, v\rangle \in E$ ( note that nodes $p_1, \ldots, p_m$ are not necessarily distinct). The set of possible values which $M(v)$ might be set to is:

$$S_P = \{h_m(p) \mid p \in P\}$$

Each time the possibility of further visiting $v$ is considered, $M(v)$ is compared with a value $s \in S_P$ and the node is further enqueued in $Q$ iff $M(v)$ is strictly lower than $s$. The proof follows by considering that every time $v$ is visited, the value of $M(v)$ strictly increases, and the set $S_P$ is finite. $\square$

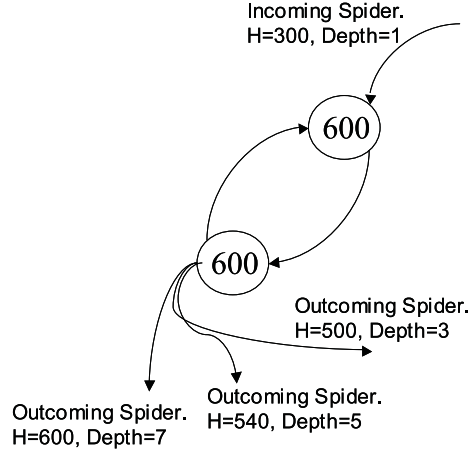

**Figure 7. A circumstance in which allowing multiple visits to a document leads to unwanted happiness levels**

**Theorem 4.14** *It is given a webgraph $G = \langle N, E, r \rangle$, a starting node $u \in N$, an ending node $v \in N$, a rational $ht$ and an integer $m$. If there exists a promising path between $u$ and $v$, Algorithm 4.12 will reach $v$ when started from $u$.*

**Proof.** (Sketch). The proof is carried out by induction on the length of the promising path between $u$ and $v$. If $u$ and $v$ are connected through the trivial promising path $\langle u, v \rangle$, clearly, Algorithm 4.12 will visit $v$. Furthermore, we can conclude that $M(v)$ is set to a final value greater or equal than $h_m(\langle u, v \rangle)$.

Assume $v$ is connected to $u$ through a promising path $p = \langle u, p_1, \ldots, p_h, v \rangle$. The path $p' = \langle u, p_1, \ldots, p_h \rangle$ is obviously promising.

We assume Theorem holds for nodes $u$ (starting node) and $p_h$ (ending node), and, furthermore, that $M(p_h)$ is set to a final value greater or equal than

$$T_{p_h} = h_m(\langle p_{max(0,h-m)}, \ldots, p_{h-1}, p_h \rangle)$$

where $p_0 = u$. Thus, $p_h$ must have been enqueued at most once in $Q$ through a pair $\langle \langle s_1, \ldots, s_t \rangle, p_h \rangle$ such that $h_m(\langle s_1, \ldots, s_t \rangle) \geq T_{p_h}$ and $T_{p_h}$ has to be greater than $ht$: should this not be true, $p'$ would not be a promising path.

This enforces $v$ to be visited also. Furthermore, we have to conclude that the final value of $M(v)$ is greater or equal than

$$T_v = h_m(\langle p_{max(0,h+1-m)}, \ldots, p_h, v \rangle)$$

where $p_0 = u$. □

Although we have shown termination of evaluation process is guaranteed (independently from the maximum depth constraint), in GSA, we preferred not to adopt Algorithm 4.12 since this might induce artificially "happy" spiders.

> Input: a set of good documents *G*, a set of bad documents *B*, and a set of keywords *W*;
> Output: a set of good words *G*;
>
> **begin**
> (1)  Extract the set $P$ of $k$ nearest non-noise words to any word of *W*, within $G$;
> (2)  Return the $k'$ words of $C$ with best discriminating power w.r.t. $G$ and $B$;
> **end**;

**Figure 8.   Algorithm which controls the Feedback Engine**

Figure 7 shows a counterexample where happiness of generated spiders tends to altered values, since nodes forms cycles. Figure shows two mutually linked nodes each one with $r$ value set to $600$. An incoming spider with initial happiness set to $300$ would yield many generations of spiders induced by multiple loops on the two nodes (we assume that happiness is computed on the last five visited nodes, that the incoming spider has visited just one node, and that $r$ values range between $0$ and $1000$).

Furthermore, we argue that this approach, equipped with suitable cycle control algorithms (e.g. allowing to visit only those paths with no duplicate nodes), would lead to slight improvements in recall performances, with respect to speed performance decay (indeed, consider that each node can be visited, in the worst case, $O((|N|m)^m)$ times).

## 5  Exploiting user preferences

GSA also features some user profiling technique, intended to enrich the system with useful information on what and on how to perform searching. User may express her/his preferences in two ways:

1. by manually editing GSA concept tree, in order to better drive anticipated exploration. Role of concept tree was detailed in Section 4.

2. by expressing preferences on retrieved documents. As we will see in the present Section, GSA is able to analyze such preferences producing sets of interesting words, which can be fruitfully used by users to suitably refine queries and concepts.

Following [25], the user can express a boolean preference about each of the retrieved documents, or ignore some of them (i.e. a document might be marked either as "hot" or "cold"). She/he can then ask GSA to take care of the preferences she/he expressed. GSA examines hot and cold documents and produces a set of *good* terms and a set of *bad terms* (the latter are not considered in the current release of the system), which are presented to the user. The user is then able to handle these terms in order to refine GSA's concept tree.

The algorithm followed by Feedback Engine is shown in Figure 8. The input of the Algorithm is a set of words $W$ (a query), a set of *good* documents $G$ and a set of *bad* documents $B$, whereas the output of the Algorithm is a set of *good* words. It basically consists of two different steps:

1. A pre-selection is made by extracting only $k$ candidate words. Candidate words are selected by considering those words which have at least one occurrence close to (i.e. with a distance within a given number of words) an occurrence of a word of $W$, within documents of $G$. A set of *noise words* [9], containing very common terms, is a priori excluded from this computation. Words occurring very near to words of $W$ are preferred with respect to farther ones.

2. Each candidate word is scored according to a suitable measure of interestingness (the discriminating power), and the $k'$ best words are output.

Intuitively, a good word is such that interesting pages take definitely a higher score w.r.t. uninteresting documents, when this word is applied as search keyword. Thus, we replaced the analogous measure of expected information gain of [25], with a comparable measure of *discriminating power*. In other terms, discriminating power is employed in order to estimate how much a candidate word would be effective if applied as query word joint with other words of $W$.

**Definition 5.1** Let $G$ be a set of good documents and $B$ be a set of bad documents; let $W$ be the set of words of the original query. The *discriminating power* $dp$ of a term $t$ is defined as:

$$dp(t, G, B) \;=\; \sum_{g \in G} \frac{rank(g, W \cup \{t\})}{|G|} - \sum_{b \in B} \frac{rank(g, W \cup \{t\})}{|B|}$$

Each term increments its discriminating power if it obtains a high score in a good document and decrements its discriminating power if it contributes to enhance $rank$ value in a bad document. Since the distance component is quite influencing, this technique naturally privileges those terms which occur near to the original query words.

This approach is in some sense similar to traditional Bayesian classification applied to text documents [17, 25], where each word is considered as a (boolean) attribute, and the task of classifying new documents as "relevant" or "not relevant" is done by estimating how much is probable that a given document belongs to one of the two classes.

However we chose not to adopt usual classification methods: in order to be effective, such techniques leave an heavy burden on the user, requiring to classify large sets of documents. Thus, we preferred a good heuristic technique, which showed interesting performances, and requires to handle relatively small sets of documents.

## 6 An example session

A typical session screenshot of GSA is shown in Figure 9. The screen is divided into two main sections: a concept editing frame, and a document view frame. On the left-hand side it is possible to edit the concept tree by adding, removing and modifying concepts (folder-shaped icons) and queries (sheet-shaped icons). By dragging and dropping nodes it is possible to change the tree structure. For example, clicking on the apposite button, the user can enter a query (e.g. "Voyager [...]"), and then drag & drop the corresponding icon under the chosen concept icon (e.g. "Nasa", under the more general entry "Space Exploration"). Query's keywords (in this case, "voyager" and "project") are displayed on the top of the concept tree frame, whenever the user clicks the corresponding item.

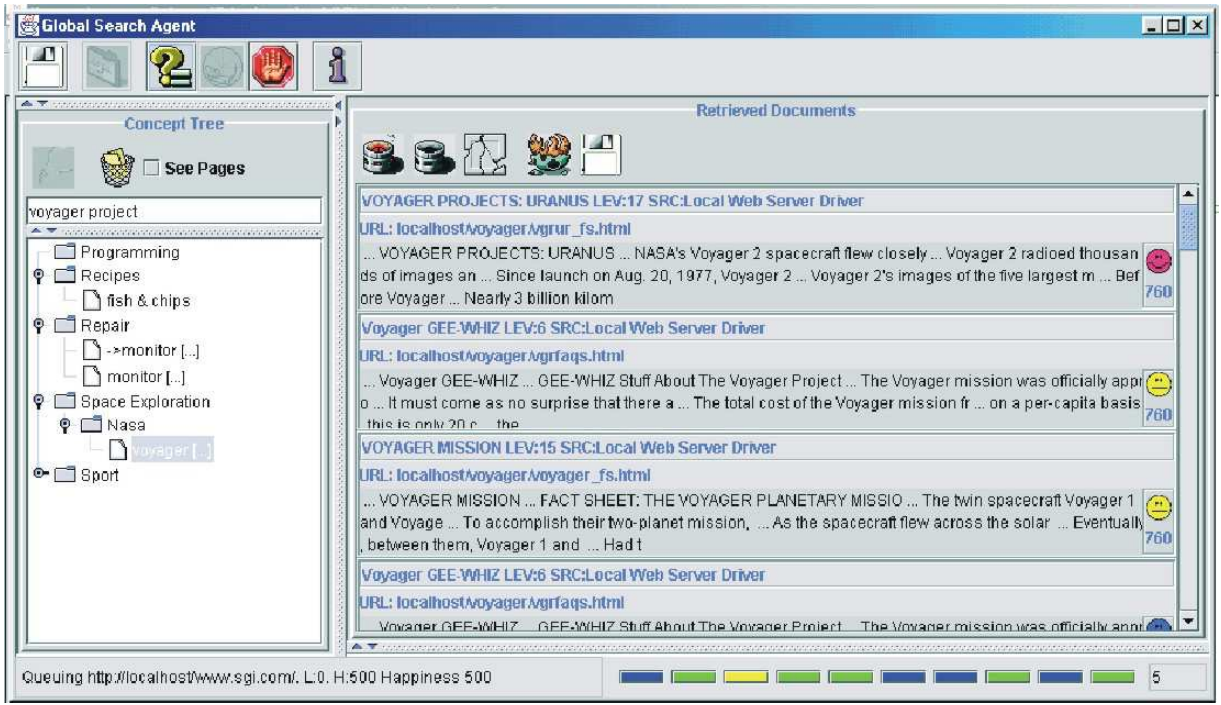

**Figure 9. A typical GSA session**

Documents retrieved for the currently selected query are displayed on the right-hand side. Each document entry reports:

- the level of depth of the spider that found the item,
- the name of the search engine originating the generation of spiders that found the item (where level 0 is associated to documents directly extracted from some search engine),
- the user rating of the document, where smiling faces represent items marked as good and sad faces are used to refer to bad ones.
- a page title, a short abstract and the overall score.

The search process can be activated by selecting a query and acting on the apposite button. Figure 9 reports system activity short after the activation of the query "Voyager [...]". The document view frame is populated as soon as there are results to be displayed.

An array of colored frames on the bottom of the page reports the activity of running spiders. The color of a frame indicates the status of a corresponding spider ( e.g. "waiting for connection", "connected", "parsing" etc. ). Tones of blue are proportional to spider's happiness, where lighter tones correspond to spiders with higher happiness.

The overall number of spiders waiting for activation is shown on the bottom-left of the screen. The user can stop the search process in any moment.

After marking documents as good (e.g. the topmost document in Figure 9) or bad, the Feedback Engine can be started, by acting on the apposite button.

The Feedback Engine analyzes documents, and, as soon as the derivation process terminates, a new query is popped up within the concept tree (e.g. "→monitor [...]"). Derived queries are enriched with suggested words (good words), which can be also edited by the user.

Other features can be exploited, which are not illustrated in this paper, such as scheduling periodic queries, the possibility to push and pop query requests to and from remote instances of the agent and the possibility to save and display document contents within the application or using preferred browsers.

# 7 Experimental results

In order to prove effectiveness of our approach to searching, we thoroughly tested the current prototype in some different contexts, where evaluation scenarios are pretty different. The first two test sets of queries are used to verify spider effectiveness within two different environments where the locality principium holds to a precise extent, whereas the third test set of queries was conducted in order to appreciate general GSA performances.

1. The first set of tests (test set 1) was conducted off-line, exploring a set of 24'038 documents for a total of about 400MB of text information. Documents were collected from a set of 87 sites snapshots, regarding heterogenous topics. Although quantity of documents for this test set is reasonably huge, the size of the explored webgraph is clearly definite. Thus tests conducted on these documents allowed to appreciate time savings, in terms of percentage of disregarded documents (recall GSA tends to explore only "promising" paths). Data gathered during tests allowed to estimate to what extent the locality principium holds in this document set.

   Figure 10 and Table 2 report aggregate values for $P(rank(X) = t \mid rank(Y) = s)$, i.e. probability that a document $X$ scores a $rank$ value $t$ whenever it is known that it is referred by a document $Y$, scoring a $rank$ value of $s$.

   As can be seen, on 43'969 evaluations of $rank$ (taken on many different queries), only $2.6\%$ of them scores a $rank$ value higher than 700. However, focusing just on documents reached from nodes with $rank$ value better than 700, this probability boosts to $40.4\%$.

   The overall correlation value between $rank(X)$ and $rank(Y)$ was found to be 0.54.

2. The second set of tests (test set 2) was conducted on-line, by starting GSA spiders on Google results, in order to test improvement given by GSA to Google hits. Table 3 and Figure 11 reports aggregate values for $P(rank(X) = t \mid rank(Y) = s)$ in this setting. Data was collected on a set of $6521$ $rank$ evaluations. It can be noted that the overall quality of documents is, in this case, very high ( $P(rank(X) > 700) = 22.34\%$ ). Locality principium can be appreciated on lower values (e.g.

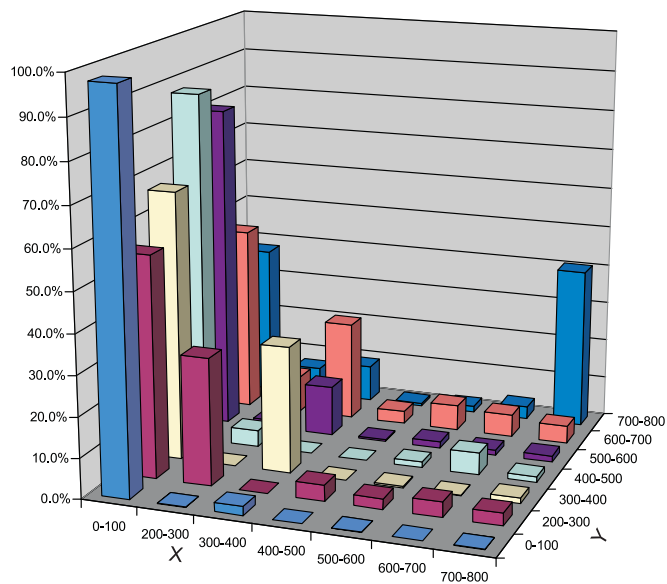

**Figure 10. Histogram of $P(rank(X) = t | rank(Y) = s)$ where $Y$ is a document linking $X$ for the test set 1.**

$P(rank(X) < 100 \mid 300 \leq rank(Y) \leq 400)$ is 100%). Note that there were not enough data to estimate $P(rank(X) = t \mid rank(Y) < 100)$.

3. The third set of tests (test set 3) was conducted by comparing GSA with Altavista, Excite, Google and Hotbot, in order to test overall performance of GSA as a metasearch engine, in a fixed query duration setting.

It is worth noting that $rank$ tends to cluster documents into three distinct sets (documents scoring less than 100, documents scoring from 300 to 400, and documents scoring more than 700). This is mainly due to the granularity of $rank$ with respect to the distance and presence components, which become relevant whenever one or more query terms are actually present in a document. In fact, $rank$ tends to score values higher than 700 whenever all the query terms are present in a document (usually a query contains two terms), and a value between 300 and 400 when one term is present.

The connection speed was about 2Mbit/sec. Constants chosen for the scoring function were $k_0 = 1000, k_1 = 10, k_2 = 10, k_3 = 1, k_4 = 20, k_5 = 250, k_6 = 300$. Happiness was computed on the last five visited documents.

### 7.1 Tests conducted off-line

The scenario of test set 1 reflects typical circumstances in which GSA is employed as search engine on unexplored, unindexed sites. Since, in this context, percentage of relevant documents is very low, in order to induce an "optimistic" spider behavior, test set 1 was performed by setting initial happiness to 500 and happiness threshold to 251, while no constraints on the maximum reachable depth were introduced: thus GSA was made able to potentially explore the whole document set. We submitted eleven queries whose results are accounted for in Table 4. For each query, we present values of *saving* and *recall*. We introduced a measure of "saving" $s$,

$$s = \frac{|L| - |E|}{|L|}$$

where $E$ is the set of explored documents during a given query evaluation, and $L$ is the experimental document set. Saving gives evidence of the fraction of ignored documents w.r.t. the entire document set, and thus it is a direct measure of the reduction for the search space explored during the search.

|         | rank(X)  |         |         |         |         |         |         |        |
|---------|----------|---------|---------|---------|---------|---------|---------|--------|
| rank(Y) | 0-100    | 200-300 | 300-400 | 400-500 | 500-600 | 600-700 | 700-800 | Total  |
| 0-100   | 97.6%    | 0.1%    | 2.2%    | 0.0%    | 0.0%    | 0.0%    | 0.1%    | 100%   |
| 200-300 | 55.0%    | 31.6%   | 0.0%    | 4.0%    | 2.6%    | 3.7%    | 3.1%    | 100%   |
| 300-400 | 66.8%    | 0.0%    | 31.5%   | 0.0%    | 0.4%    | 0.1%    | 1.3%    | 100%   |
| 400-500 | 87.7%    | 4.1%    | 0.0%    | 0.0%    | 1.4%    | 5.5%    | 1.4%    | 100%   |
| 500-600 | 80.9%    | 1.8%    | 12.5%   | 0.4%    | 1.6%    | 1.2%    | 1.5%    | 100%   |
| 600-700 | 46.6%    | 9.2%    | 24.7%   | 3.2%    | 6.3%    | 5.7%    | 4.3%    | 100%   |
| 700-800 | 38.0%    | 7.1%    | 9.1%    | 0.7%    | 1.6%    | 3.1%    | 40.4%   | 100%   |
| Total   | 85.0%    | 1.4%    | 8.5%    | 0.2%    | 0.3%    | 0.4%    | 2.6%    | 100%   |

**Table 2. Tabular values for $P(rank(X) = t \mid rank(Y) = s)$ where $Y$ is a document linking $X$ for the test set 1.**

|         | rank(X)  |         |         |         |         |         |         |        |
|---------|----------|---------|---------|---------|---------|---------|---------|--------|
| rank(Y) | 0-100    | 200-300 | 300-400 | 400-500 | 500-600 | 600-700 | 700-800 | Total  |
| 300-400 | 100.0%   | 0.0%    | 0.0%    | 0.0%    | 0.0%    | 0.0%    | 0.0%    | 100%   |
| 500-600 | 66.7%    | 0.0%    | 33.3%   | 0.0%    | 0.0%    | 0.0%    | 0.0%    | 100%   |
| 600-700 | 55.6%    | 7.8%    | 10.0%   | 0.0%    | 0.0%    | 22.2%   | 4.4%    | 100%   |
| 700-800 | 34.6%    | 11.1%   | 29.3%   | 0.3%    | 0.7%    | 1.4%    | 22.6%   | 100%   |
| Total   | 34.9%    | 11.0%   | 29.0%   | 0.3%    | 0.7%    | 1.7%    | 22.3%   | 100%   |

**Table 3. Tabular values for $P(rank(X) = t \mid rank(Y) = s)$ where $Y$ is a document linking $X$ for the test set 2.**

Let $L$ be the experimental document set, and $R$ be the set of ten documents with highest $rank$ value within $L$. Then, recall $r$ is defined as:

$$\frac{|F \cap R|}{|R|}$$

where $F$ is the set of documents found and presented to the user by GSA for a given query.

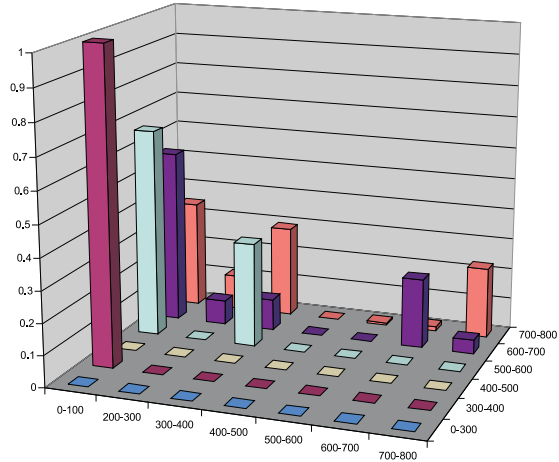

**Figure 11. Histogram of $P(rank(X) = t | rank(Y) = s)$ where $Y$ is a document linking $X$ for the test set 2.**

Recall allows to appreciate how many interesting documents were ignored, due to the search space reduction introduced by GSA. Objective quality of a document is assumed to be its $rank$ value.

| Query # | Keywords | Explored documents | Saving | Recall |
|---|---|---|---|---|
| 1 | javascript tutorial | 7328 | 69.51% | 100% |
| 2 | sailing course | 2900 | 87.94% | 100% |
| 3 | jubilee calendar | 854 | 96.45% | 30% |
| 4 | cindy crawford | 149 | 99.38% | 100% |
| 5 | neural networks | 293 | 98.78% | 100% |
| 6 | english courses | 4208 | 82.49% | 70% |
| 7 | monitor self repair | 1768 | 92.64% | 80% |
| 8 | self woodworking | 528 | 97.80% | 100% |
| 9 | tama snare drums | 381 | 98.42% | 100% |
| 10 | old computers | 305 | 98.73% | 30% |
| 11 | fast dimm memory | 203 | 99.16% | 20% |
|  | **Avg** | **1719** | **92.76%** | **75.45%** |

**Table 4. Results for test set 1**

Saving values are very high, and give evidence of a surprising reduction of the search space. In six cases out of eleven, high savings values did not compromised recall at all. It is worth to consider that some low recall values obtained during tests are due to the $rank$ function accuracy.

In fact, for queries $3, 10$ and $11$ of Table 4, we noticed that $R$ (the set of ten documents with highest score) contained many uninteresting (from the user-perceived point of view) documents. In these cases, the set $F$ (best documents retrieved by GSA), was perceived by users as highly valuable instead. So, it can be conjectured that a $rank$ function taking in account other factors, such as quality of neighbor documents might be effective in order to improve performances in this respect (GSA does so implicitly, by driving spiders using happiness), as some studies by Kleinberg also indicate [18].

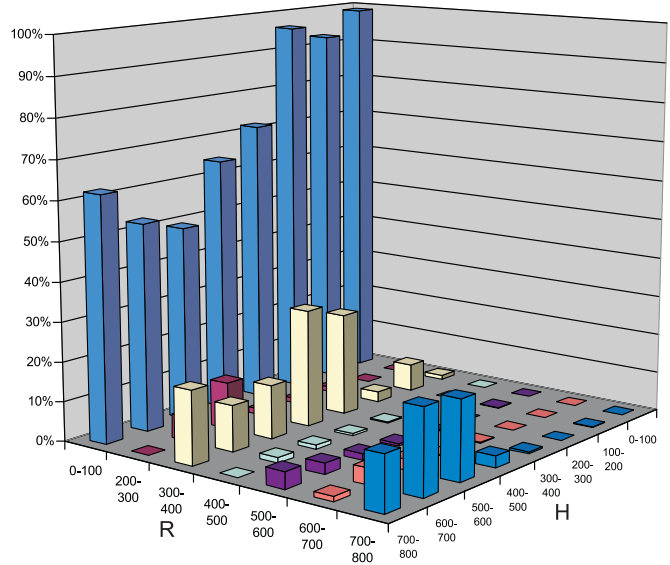

**Figure 12. Histogram of** $P(rank(X) = t \mid hap(X) = s)$ **where** $X$ **is a document reached with happiness** $s$.

In order to put into evidence the dependency between happiness and quality of documents reached we estimated the probability $P(rank(X) = t \mid hap(X) = s)$, i.e. the probability that document $X$ has $rank$ value $t$ whenever it is reached by a spider with happiness $s$. Results are given in Figure 12 and Table 5. The overall correlation value between $rank(X)$ and $hap(X)$ was found to be $0.47$.

### 7.2 Tests conducted on-line

Test set 2 reproduces a scenario in which GSA is employed as accompanying tool for Google. Since the expected quality and quantity of reached documents is high in this setting (see Table 3), GSA was set up in a "pessimistic" fashion, with a happiness threshold of 700, a starting happiness of 0, and a maximum depth threshold of 2. In spite of this latter constraint, GSA was able to reach a considerable quantity of documents.

Table 6 report results obtained on the nine queries constituting test set 2. The rightmost column accounts for results produced directly by GSA within upon the ten most relevant documents. In particular, let *improvement* be defined as:

$$i = \frac{\sum_{i>1} |F_i|}{10}$$

|         | **rank(.)** |         |         |         |         |         |         |         |
|---------|---------|---------|---------|---------|---------|---------|---------|---------|
| **hap(.)** | 0-100 | 200-300 | 300-400 | 400-500 | 500-600 | 600-700 | 700-800 | Total |
| 0-100   | 99.0%   | 0.0%    | 1.0%    | 0.0%    | 0.0%    | 0.0%    | 0.0%    | 100%    |
| 100-200 | 92.8%   | 0.0%    | 6.9%    | 0.0%    | 0.1%    | 0.0%    | 0.1%    | 100%    |
| 200-300 | 96.2%   | 1.2%    | 2.6%    | 0.0%    | 0.0%    | 0.0%    | 0.0%    | 100%    |
| 300-400 | 71.7%   | 1.2%    | 25.8%   | 0.2%    | 0.3%    | 0.2%    | 0.4%    | 100%    |
| 400-500 | 64.5%   | 1.5%    | 29.6%   | 0.6%    | 0.6%    | 0.3%    | 3.0%    | 100%    |
| 500-600 | 49.4%   | 11.7%   | 13.5%   | 1.0%    | 1.6%    | 2.7%    | 20.0%   | 100%    |
| 600-700 | 52.7%   | 6.1%    | 11.6%   | 1.4%    | 2.8%    | 4.1%    | 21.4%   | 100%    |
| 700-800 | 62.0%   | 0.0%    | 18.5%   | 0.0%    | 4.3%    | 1.3%    | 13.9%   | 100%    |
| **Total** | **86.3%** | **1.5%** | **8.6%** | **0.2%** | **0.4%** | **0.4%** | **2.6%** | **100%** |

**Table 5. Tabular values of** $P(rank(X) = t \mid hap(X) = s)$**, where** $X$ **is a document reached with happiness** $s$**, for test set 1.**

| Query # | Keywords | Explored documents | Duration (sec) | Improvement |
|---------|----------|--------------------|-----------------|-------------|
| 1 | jubilee calendar | 70 | 60 | 40% |
| 2 | sailing course | 2906 | 2500 | 50% |
| 3 | english courses | 468 | 392 | 60% |
| 4 | self woodworking | 517 | 331 | 70% |
| 5 | dimm memory pinout | 88 | 301 | 50% |
| 6 | lisp tutorial | 1982 | 2089 | 20% |
| 7 | neural networks faq | 2744 | 1833 | 50% |
| 8 | tama snare drums | 1379 | 677 | 40% |
| 9 | cindy crawford | 1216 | 379 | 50% |
|   | **Avg** | **1263** | **951** | **47.7%** |

**Table 6. Experimental Results for test set 2.**

where $F_i$ is the set of documents retrieved by GSA using a spider at depth level $i$ (documents directly retrieved from Google are at level 0), and which are placed within the ten most relevant documents.

It can be noted that, in the average, GSA enriches relevant pages produced by Google with about the same quantity of pages of the same quality. Although results are promising, current GSA implementation limits did not allowed relevant performances in term of speed of evaluation.

Test set 3 reproduces conditions in which GSA is employed as metasearch engine. In this case we appreciated performances of GSA on fixed duration queries. Table 7 reports results on the set of 20 queries constituting test set 3. For each query we considered the ten most relevant documents reported from each search engine. The table indicates, for each search engine, the overall number of documents found, the (user-perceived) number of documents considered as relevant, and the precision (percentage of relevant documents

|  | # Documents Found | # Relevant documents | Precision |
|---|---|---|---|
| GSA (30 sec.) | 174 | 103 | 59.2% |
| Altavista | 200 | 75 | 37.5% |
| Excite | 200 | 82 | 41.0% |
| Google | 200 | 166 | 83.0% |
| Hotbot | 200 | 48 | 24.0% |

**Table 7. The results of GSA versus some other well known search engine (test set 3).**

w.r.t. retrieved documents). Results from GSA were obtained by stopping the evaluation after 30 seconds.

Usually, GSA performs fast and very well against single search engines when short duration searches are submitted; the overhead taken by the task of directly retrieving each document is, in this setting, far balanced when five or more search engines are queried simultaneously (tests were performed doing parallel meta-search on Altavista, Excite, FastSearch, Google, Hotbot, Lycos, FastSearch, Yahoo, and Webcrawler [1, 7, 8, 11, 13, 8, 28, 27]). In this test set, Google outperformed GSA. As shown during test set 2 evaluation, GSA outperforms Google if no thresholds on the maximum evaluation time are given.

## 8. Conclusions

In this paper we illustrated GSA, a web search agent, which features meta-searching, profile learning, and, especially, anticipated and selective exploration of the Web.

Tests put into evidence effectiveness of the anticipated and selective exploration approach in terms of documents saved from exploration, and improvement of the overall quality of retrieved documents, with respect to traditional search engines.

It might be argued that the choice of explicitly retrieving and ranking documents may compromise efficiency. Indeed, the current prototype, entirely written in Java, performs badly in terms of absolute throughput (it is not able to take full advantage of high bandwidths: despite many tests were conducted on the Internet using a bandwidth of 2Mbits/sec, system parsed documents at a speed of about 50Kbits/sec).

However, this approach is justified in several application contexts, where documents have to be necessarily retrieved or where unavoidable payloads are softened by clear advantages. Possible jobs where the technique might be successfully applied are:

- Periodic server-side collection of specific topic documents (e.g. user-profile compliant news) to be pushed to the user attention;

- Smart and rapid indexing of new web sites, whenever indices size has to be kept small and indices are to be produced quickly;

- Agent assisted browsing, featuring intelligent and anticipated exploration.
- Highly parallelized meta-searching. In this case, payloads are significantly blurred by parallel retrieval of many documents.

At the moment we are studying several improvements to the architecture of GSA, like the automatic generation of engine-dependent wrappers [24, 23]. We argue spider behavior could be improved by studying more sophisticated happiness functions and/or introducing logic based behaviors. Nonetheless, it is worth to reconsider ranking function formulation, in terms of "originality" of the proposed document (e.g. it should be avoided that many, very similar, documents from the same site are presented to the user).

Although, at the moment, concepts trees are completely managed by users, we believe the system could be improved by introducing an automated concept tree derivation like in [29], and providing an automated parameter tuning [2].

## Acknowledgements

The author wishes to thank Antonio Castellucci, Domenico Vasile and Sebastiano Costa, for several thoughtful discussions on the earlier development of GSA, and for participating to the software implementation and maintenance. The author also appreciated referees' comments which helped to significantly improve the quality of this paper.